\documentclass[12pt]{article}

\usepackage[margin=1in]{geometry}
\usepackage{amsmath,amsfonts,amssymb}
\usepackage{algorithmic}
\usepackage{algorithm}
\usepackage{array}
\usepackage[caption=false,font=normalsize,labelfont=sf,textfont=sf]{subfig}
\usepackage{textcomp}
\usepackage{stfloats}
\usepackage{url}
\usepackage{verbatim}
\usepackage{graphicx}
\usepackage{makecell}
\usepackage{cite}
\usepackage{multirow}
\usepackage{booktabs}
\usepackage{tabularx}
\usepackage{hyperref}
\usepackage{authblk}
\usepackage{abstract}
\usepackage{titlesec}
\usepackage{parskip}

\hypersetup{
    colorlinks=true,
    linkcolor=blue,
    citecolor=blue,
    urlcolor=blue
}

\title{\textbf{CAM-VFD: Cross-Attention Multimodal \\ Video Forgery Detection}}

\author[1]{Hoda Osama Elkhodary}
\author[1]{Sherin Mostafa Youssef}
\author[1]{Marwa Elshenawy}
\author[1]{Dalia Sobhy}

\affil[1]{Computer Engineering Department, College of Engineering and Technology,
Arab Academy for Science, Technology and Maritime Transport, Alexandria, Egypt\\
\texttt{\{hudaoelkhodary, sherin, marwa\_elshenawy, dalia.sobhi\}@aast.edu}}

\date{}

\begin{document}

\maketitle

\begin{abstract}
The rapid advancement of Deepfake technologies and video manipulation tools
poses a critical challenge to multimedia forensics, judicial evidence integrity,
and information authenticity. Current detectors rely on single-modality signals,
treating appearance, geometry, and motion independently. However, advanced
generators maintain within-modality consistency while producing cross-modal
contradictions, which are forensically discriminative but invisible to any
single-modal detector. We propose CAM-VFD, a Cross-Attention Multimodal Video
Forgery Detection framework that models cross-modal contradiction as a
directional forensic signal. The framework uses a cross-attention fusion
mechanism in which CLIP-based appearance representations serve as queries
against VideoMAE motion features and MiDaS depth features, enabling the
identification of contradictions between visual, temporal, and geometric
evidence. We examine this design through cross-modal attention discrepancy
analysis, observing statistically separable real and fake distributions
($p<0.001$, Cohen's $d=0.68$). Experimental results on two generative video
benchmarks indicate consistent performance, with 95.31\% Top-1 accuracy on
GenVidBench and 93.43\% accuracy, 90.63\% F1-score, and 96.56\% AUROC on
GenVideo. Moreover, CAM-VFD demonstrates stable performance under compression,
noise, blur, and adversarial perturbations, suggesting that cross-modal
reasoning may improve robustness in media forensics. The code is publicly
available at \url{https://github.com/Hoda-Osama/CAM-VFD/tree/main}.
\end{abstract}


\newpage

\section{Introduction}

Recent evolution in artificial intelligence, particularly deep learning-based
techniques such as Generative Adversarial Networks (GANs), diffusion models,
and auto-regressive architectures, has enabled AI-based synthetic video
generators to emerge. Deepfakes---realistic yet fabricated synthetic
videos---are becoming increasingly difficult to distinguish with the human eye,
challenging the validity of digital media in courts~\cite{bib1}. The
consequences are severe: fabricated videos have appeared as forensic evidence
in legal proceedings, been exploited to spread political disinformation, and
used to impersonate individuals without consent~\cite{bib2}. Consequently,
there is an urgent need for reliable and scalable methods for Deepfake
detection.

Previous work in video forensics has progressed from detecting spatial
artifacts in individual frames~\cite{bib3,bib4} to capturing temporal
inconsistencies in facial dynamics~\cite{bib5,bib6,bib7}. More recent
approaches have explored spatio-temporal feature learning~\cite{bib8,bib9}
and semantic coherence analysis~\cite{bib10,bib11}. However, these approaches
fail to generalize to the next-generation generators, which can now synthesize
entire semantically rich scenes rather than modifying facial regions using
generative models of text-to-video (T2V)~\cite{bib12} or image-to-video
(I2V)~\cite{bib13}. Although some recent approaches designed for full-frame
manipulations have been developed~\cite{bib14}, a fundamental limitation
persists in nearly all of them: they treat the video as a collection of
signals from a single modality dimension, overlooking cross-modal
contradictions and misalignment between appearance, geometry, and motion.
These cross-modal contradictions cause a detection failure in
current-generation AI video detection models because most AI-generated videos
may appear locally plausible in any single modality while containing
cross-modal contradictions rarely found in physically captured authentic
footage. Although multimodal detection technologies are introduced in some
recent works~\cite{bib15,bib16}, they are still incapable of combining
appearance, depth, and motion cues through effective fusion mechanisms.
Specifically, most existing multimodal frameworks are not designed to
explicitly detect cross-modal contradiction as a forensic signal; they seek
to benefit from complementary information without querying one modality for
evidence that contradicts another.

In response to these challenges, we propose \textbf{CAM-VFD} (Cross-Attention
Multimodal Video Forgery Detection), which treats cross-modal contradiction as
an explicit, \emph{directional} forensic signal. Appearance is the richest
semantic forensic channel as it integrates semantic content, texture,
illumination, and object identity, so appearance tokens serve as a forensic
query channel while motion and depth provide the key evidence against which
appearance is tested. The model computes, for each appearance token, the
degree to which the motion and depth context are consistent with that
appearance.

\noindent Our proposed framework provides the following contributions:
\begin{itemize}
  \item We introduce \textbf{cross-modal contradiction} as a directional
  forensic signal for Deepfake detection, and validate its forensic
  separability via attention discrepancy analysis. The results demonstrate a
  clear and consistent separation between real and fake samples.

  \item We propose \textbf{a cross-attention fusion module} in which
  CLIP~\cite{bib17} appearance representations serve as queries against
  VideoMAE~\cite{bib20} motion and MiDaS~\cite{bib18,bib19} depth as keys
  and values, operationalizing the directional forensic relationship.

  \item We conduct a \textbf{comprehensive evaluation on two benchmarks} to
  prove the efficacy of our proposed framework for fake video detection
  compared to state-of-the-art methods. We tested the model across
  GenVidBench and GenVideo, achieving higher accuracy compared to the
  state-of-the-art.

  \item We perform \textbf{robustness and security analysis} under real-world
  conditions, including compression, noise, blurring, and adversarial
  perturbations, with direct SOTA comparison against DeMamba under identical
  settings. The findings suggest improved robustness to challenging
  degradations while preserving reliable detection performance.
\end{itemize}

The remainder of this paper is structured as follows.
Section~\ref{sec:related} discusses the related work.
Section~\ref{sec:method} describes the proposed CAM-VFD framework.
Section~\ref{sec:experiments} presents the experimental results and ablation
studies. Section~\ref{sec:discrepancy} presents the cross-modal attention
discrepancy analysis validating our forensic premise.
Section~\ref{sec:robustness} reports robustness and adversarial evaluation.
Section~\ref{sec:conclusion} concludes the paper.

\section{Related Work}
\label{sec:related}

\begin{table}[t]
\centering
\small
\setlength{\tabcolsep}{5pt}
\renewcommand{\arraystretch}{1.1}
\caption{Summary of existing literature.}
\label{tab:related_work}
\begin{tabular}{lccc}
\toprule
\textbf{Ref.} & \textbf{Target Content} & \textbf{Modalities} & \textbf{Fusion Strategy} \\
\midrule
\cite{bib8,bib9,bib21,bib22}         & Face       & Single (Spatio-Temp.)    & None \\
\cite{bib23,bib24}                   & Face       & Single (Spatio-Temp.)    & None \\
\cite{bib5,bib10,bib11,bib25,bib26} & Face       & Single (Semantic/Motion) & None \\
\cite{bib15,bib27,bib28}             & Face       & Two                      & Simple/Static \\
\cite{bib29,bib30,bib31}             & Face/Scene & Two--Three               & Advanced/Static \\
\midrule
\textbf{Ours} & \textbf{Scene} & \textbf{Three} & \textbf{Cross-attention} \\
\bottomrule
\end{tabular}
\end{table}

In this section, we provide a discussion related to major research directions
for detecting synthetic videos: (i)~\textit{spatio-temporal inconsistencies}
(Section~\ref{sec:relatedwork_A}), which focus on defects across space and
time; (ii)~\textit{semantic and dynamic inconsistency}
(Section~\ref{sec:relatedwork_B}), which captures unnatural human motion,
action dynamics, or semantic mismatches; and (iii)~\textit{multimodal fusion}
(Section~\ref{sec:relatedwork_C}), which integrates complementary cues across
modalities to expose manipulation traces.
Table~\ref{tab:related_work} summarizes the landscape and positions CAM-VFD
relative to prior work.

\subsection{Spatio-Temporal Inconsistencies}
\label{sec:relatedwork_A}

Early detection methods analyzed video frames individually, exploiting trace
patterns left by generators in convolutional feature maps~\cite{bib3,bib4}
or frequency-domain statistics. Research then shifted to spatio-temporal
inconsistencies, motivated by the observation that temporal consistency is
difficult for generative models to maintain. For instance, early work
by~\cite{bib8} used a 3D convolutional neural network (CNN) on stacked face
regions from consecutive frames to capture unnatural transitions.
Similarly,~\cite{bib9} proposed a spatio-temporal CNN that analyzes both RGB
spatial features and temporal dynamics from optical flow maps.~\cite{bib21}
combined three ResNet-50 models targeting spatial, temporal, and structural
anomalies with fuzzy logic to handle uncertainty.~\cite{bib22} targeted
inter-frame illumination inconsistencies by proposing a learnable Illumination
Decomposition Module (IDM) that separates frames into illumination and
reflection components. Recently,~\cite{bib23} proposed a dynamic data
augmentation strategy guided by forgery heatmaps produced by a Swin
Transformer, combined with Cross-Frame Multi-Head Attention (CMA) and a
Bidirectional GRU to model sequential spatio-temporal dependencies. To
improve generalisation to unseen generation methods,~\cite{bib24} proposed
FakeSTormer, a multi-task learning framework with two auxiliary branches
targeting spatial and temporal regions independently, paired with a video
synthesis strategy redefining Deepfake detection as a fine-grained regression
task rather than a binary classification problem. Despite these advances,
spatio-temporal methods are insufficient against generators that produce
realistic frame-to-frame consistency while introducing cross-modal
contradictions invisible to within-modality analysis.

\subsection{Semantic and Dynamic Inconsistency}
\label{sec:relatedwork_B}

AI-generated videos often fail to replicate natural biomechanics and logical
scene coherence; the scene as a whole does not appear reasonable and
contextually continuous, so recent approaches have begun to explore human
motion and semantic forensics. For example,~\cite{bib5} demonstrated the
importance of facial features in this context---Deepfakes often violate
natural head motion constraints.~\cite{bib10} utilized CLIP embeddings to
find semantic mismatches between expected human behavior and generated motion
patterns. Other methods proposed in~\cite{bib25} detected unnatural temporal
flicker in synthetic movements using frequency-domain motion
representations.~\cite{bib26} proposed a cross-branch fusion to analyze
dynamic inconsistencies in facial movements.~\cite{bib11} focused on local
motion patterns between adjacent frames, which are often overlooked by
temporal models. These methods demonstrate that dynamic cues carry forensic
information, yet they operate within a single temporal or semantic dimension
without grounding evidence in geometric or appearance context.

\subsection{Multimodal Fusion}
\label{sec:relatedwork_C}

Integrating multiple modality streams has become a direction for recent
research work, motivated by the complementary forensic evidence that different
modalities provide.~\cite{bib15} introduced M2TR, a multimodal transformer
model fusing visual and spectral cues through a cross-modality fusion
block.~\cite{bib27} developed AVFF, a two-stage multimodal approach that
learns cross-modal audio-visual correspondences through self-supervised
pre-training.~\cite{bib28} proposed an Ensemble of Experts model that
synthesizes three modalities: appearance, motion, and geometry. More recent
works have substantially advanced the sophistication of multimodal fusion
strategies.~\cite{bib29} proposed CAD (Cross-Modal Alignment and
Distillation), a dual-path architecture combining a cross-modal component that
catches semantic mismatches and a distillation mechanism that preserves
important modality-specific features.~\cite{bib30} used an attention mechanism
to fuse local spatial features with global frequency-domain statistics,
demonstrating the importance of inter-modal relationships.~\cite{bib31}
proposed a multimodal Deepfake detection framework that models cross-modal
consistency between facial dynamics and scene context, demonstrating the
forensic signals in cross-modal inconsistency. The critical limitation shared
by these methods is their use of symmetric or static fusion---a fixed
aggregation of modality features---regardless of which forensic signal is most
informative for a given input. In CAM-VFD, by contrast, appearance serves as
the forensic query channel: the model asks, for each appearance token, whether
the motion and depth evidence support or contradict it.

\section{The Proposed CAM-VFD}
\label{sec:method}

In this section, we first present the overall framework of CAM-VFD, then
describe each module in detail: adaptive frame sampling, modality-specific
feature extraction, cross-modal attention fusion, and the classification
layer.

\subsection{Overview}
\label{ssec:overview}

\begin{figure*}[t]
\centering
\includegraphics[scale=0.2]{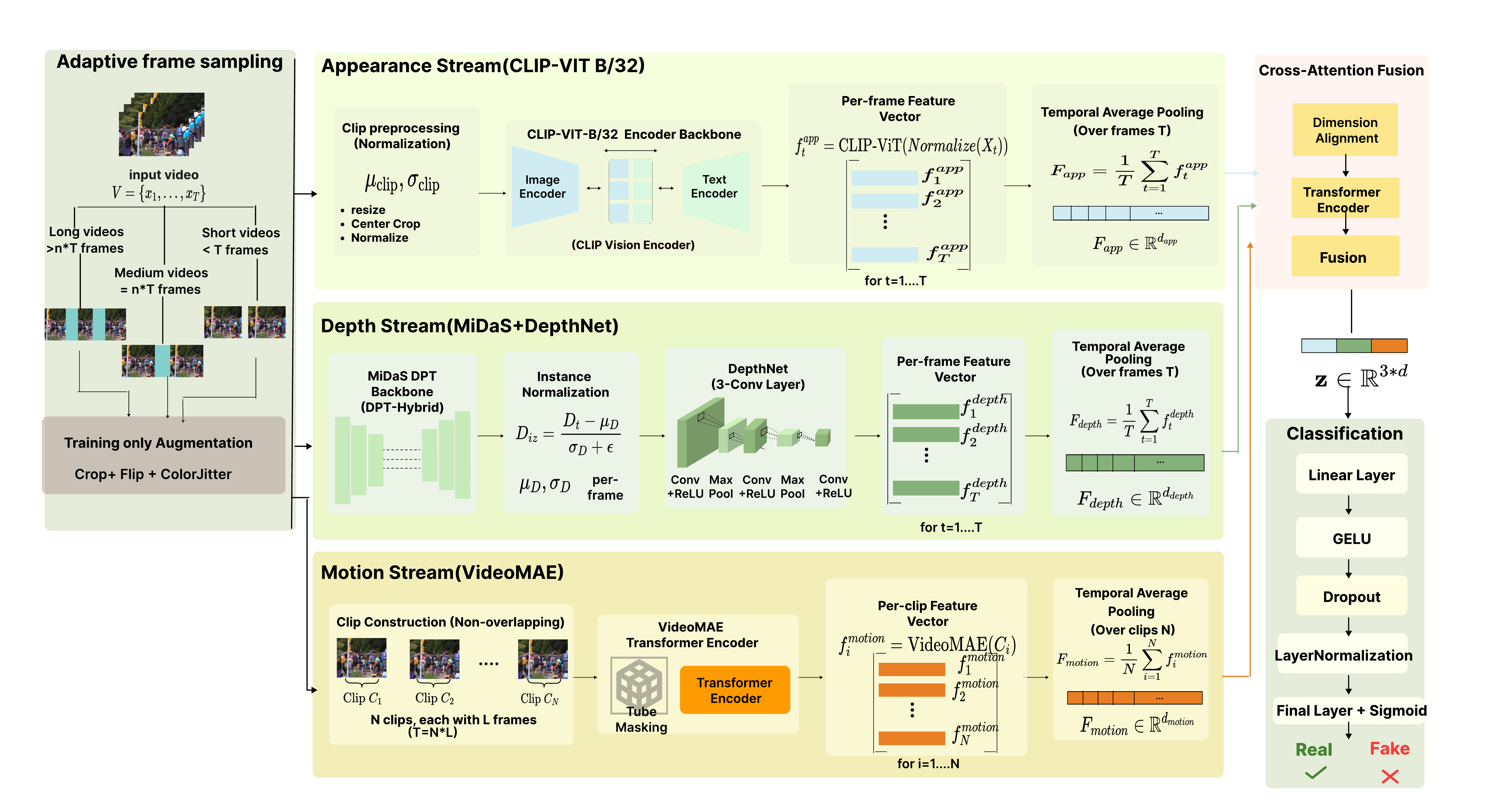}
\caption{Proposed Cross-Attention Multimodal Video Forgery Detection
(CAM-VFD) Framework.}
\label{fig:model}
\end{figure*}

The proposed CAM-VFD framework, illustrated in Figure~\ref{fig:model}, uses
three modality-specific encoders and a cross-attention fusion module operating
in a unified latent space. Given a video $\mathbf{V} = \{x_1, \ldots, x_T\}$,
the framework outputs a binary prediction $\hat{y} \in \{\text{Real},
\text{Fake}\}$.

The processing pipeline proceeds as follows. Frames are first adaptively
sampled to $T$ and augmented during training. Three pre-trained encoders then
extract modality-specific features independently: CLIP~\cite{bib17} for
appearance, MiDaS~\cite{bib18,bib19} DPT-Hybrid for depth, and
VideoMAE~\cite{bib20} for motion. All features are projected into a shared
dimensional space. Appearance features are temporally encoded by a Transformer
encoder; they then serve as queries in two parallel cross-attention blocks,
one against \textit{motion} features and one against \textit{depth} features.
The three pooled feature streams are concatenated and classified by an MLP.
Algorithm~\ref{alg:camvfd} summarizes the complete pipeline.

\begin{algorithm}[t]
\caption{CAM-VFD: Cross-Attention Multimodal Video Forgery Detection}
\label{alg:camvfd}
\footnotesize
\begin{algorithmic}
\STATE {\textbf{INPUT:}} Video $V = \{x_1, \dots, x_T\}$
\STATE {\textbf{OUTPUT:}} $y \in \{\text{Real}, \text{Fake}\}$

\STATE \textbf{1. Preprocessing:} Adaptive frame sampling + augmentations

\STATE \textbf{2. Appearance Features:}
\FOR{each frame $x_t$}
    \STATE $f_t^{\text{app}} =
           \text{CLIP-ViT}\!\left(\frac{x_t-\mu_{\text{clip}}}{\sigma_{\text{clip}}}\right)$
\ENDFOR
\STATE $F_{app} = \frac{1}{T} \sum_t f_t^{app}$

\STATE \textbf{3. Depth Features:}
\FOR{each frame $x_t$}
    \STATE $D_t = \mathrm{MiDaS}(x_t)$;\;
    $f_t^{depth} = \mathrm{DepthNet}(\mathrm{BN}(\frac{D_t - \mu_D}{\sigma_D + \epsilon}))$
\ENDFOR
\STATE $F_{depth} = \frac{1}{T} \sum_t f_t^{depth}$

\STATE \textbf{4. Motion Features:}
\FOR{each clip $C_c$}
    \STATE $f_i^{motion} = \mathrm{VideoMAE}(C_i)$
\ENDFOR
\STATE $F_{motion} = \frac{1}{N} \sum_{i=1}^{N}f_i^{motion}$

\STATE \textbf{5. Fusion \& Classification:}
\STATE $H_{app} = W_{app} F_{app}$,\; $H_{depth} = W_{depth} F_{depth}$,\;
       $H_{motion} = W_{motion} F_{motion}$
\STATE $H_{app}^{temp} = \mathrm{TransformerEncoder}(H_{app}+PE)$
\STATE $H_{app-motion} = \mathrm{Attn}(H_{app}^{temp}, H_{motion})$,\;
       $H_{app-depth} = \mathrm{Attn}(H_{app}^{temp}, H_{depth})$
\STATE $z = [\frac{1}{T}\sum H_{app}^{temp};\; \frac{1}{T}\sum H_{app-motion};\;
       \frac{1}{T}\sum H_{app-depth}]$
\STATE $p = \sigma(\mathrm{MLP}(z))$
\STATE $y = \text{Real if } p \ge 0.5 \text{ else Fake}$
\STATE \textbf{return} $y$
\end{algorithmic}
\end{algorithm}

\subsection{Adaptive Frame Sampling}
\label{subsec:sampling}

Fake videos vary substantially in length and frame rate ($fps$), which can
disrupt feature consistency and forensic traces. Fixed-interval sampling may
miss key artifact locations, and we need the most temporal coverage with
motion cues of the videos; therefore, we use an \textit{adaptive consecutive}
frame sampling technique motivated by the principle that adaptive selection
improves temporal coverage and efficiency~\cite{bib32}. We select $T$ frames
using a video-length-dependent policy: for \textit{short} videos with fewer
frames than $T$, we use cyclic repetition to preserve motion continuity. For
\textit{medium-length} videos containing $nT$ frames, we select consecutive
frame segments from different regions---from the beginning and end---to
capture both local motion smoothness and global temporal variation. For
\textit{long} videos with more than $nT$ frames, we extract distributed
segments from the beginning, middle, and end to provide broader temporal
coverage. This adaptive segmentation enables the model to detect temporal
irregularities and motion inconsistencies that are characteristic of
manipulated content.

Deep models tend to memorize pixel patterns, lighting, and other artifacts
from the training data; thus, during training only, frames are augmented with
\texttt{RandomResizedCrop} and \texttt{Flip} to simulate different viewing
conditions such as different zoom levels and camera framing, and
\texttt{ColorJitter} to enhance the model's resilience to changes in lighting
and color~\cite{bib33}.

\subsection{Modality-Specific Feature Extraction}
\label{subsec:features}

Despite the significant advancements in AI video generators, they fail to
produce fake videos that are indistinguishable from real ones. In some cases,
each appearance, motion, and depth representation may appear independently
plausible, while there is a contradiction between how a scene looks, how it
moves, and how it is geometrically structured. As illustrated in
Figure~\ref{fig:heatmap_visual}, real videos exhibit high internal consistency
across appearance, depth, and motion, whereas fake videos expose distinctive
inconsistency signatures across all three modalities. Three pre-trained
backbones extract modality-specific forensic cues independently prior to
fusion, allowing for a comprehensive analysis of the inconsistencies present
in the appearance, motion, and depth representations of the videos.

\subsubsection{Appearance Features}

Manipulated videos often contain visual attributes detectable by deep models
but not by the human eye. Common types include:
\textbf{(i) Texture Abnormalities:} unnatural skin texture, over-smoothed
textures, and inconsistent reflections and shadows;
\textbf{(ii) Boundary Artifacts:} distorted edges, misaligned facial features,
and loss of fine structural details;
\textbf{(iii) Inconsistent Illumination:} color mismatches, unnatural
gradients, and spatially inconsistent lighting.

\begin{figure}[t]
\centering
\setlength{\tabcolsep}{1.2pt}
\renewcommand{\arraystretch}{1.0}
\footnotesize
\begin{tabular}{cccc}
\multicolumn{2}{c}{\textbf{Real Video}} & \multicolumn{2}{c}{\textbf{AI-Generated Video}} \\[-1mm]
\includegraphics[scale=0.13]{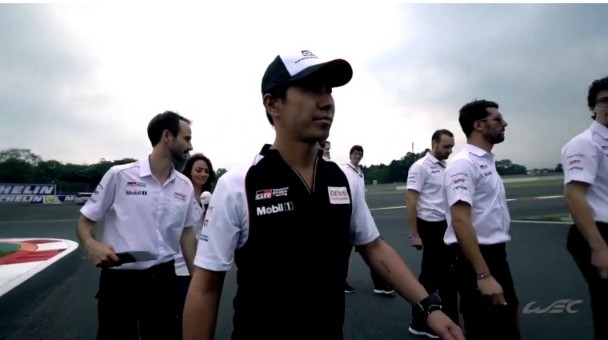} &
\includegraphics[scale=0.13]{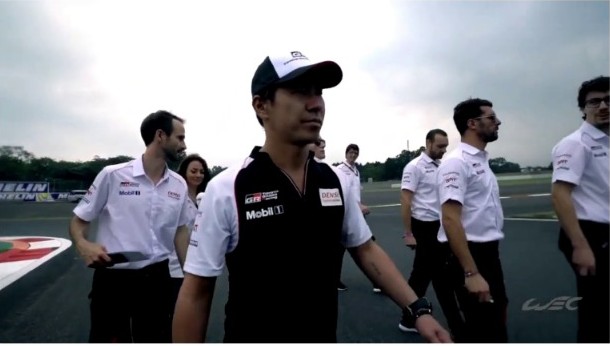} &
\includegraphics[scale=0.13]{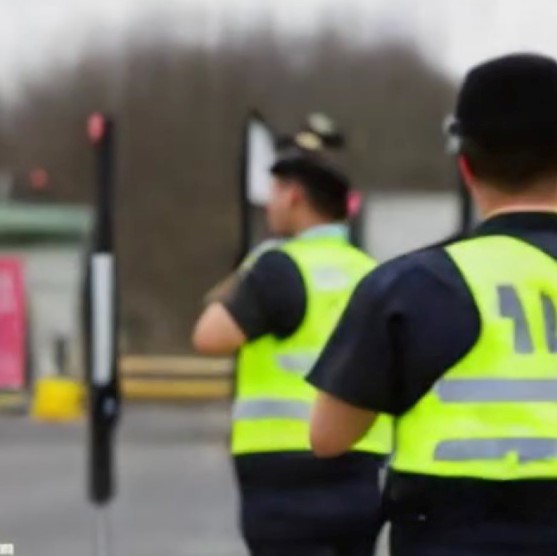} &
\includegraphics[scale=0.13]{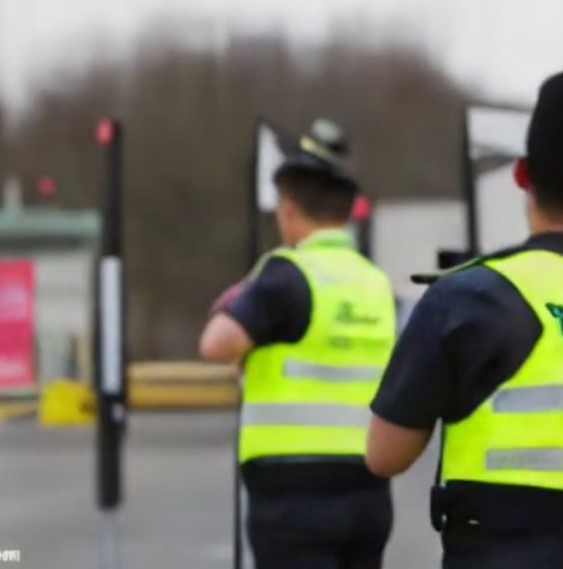} \\
\multicolumn{4}{c}{\scriptsize Sampled frames} \\[0.5mm]
\includegraphics[scale=0.13]{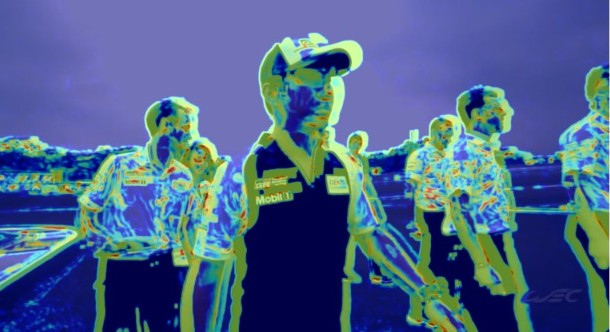} &
\includegraphics[scale=0.13]{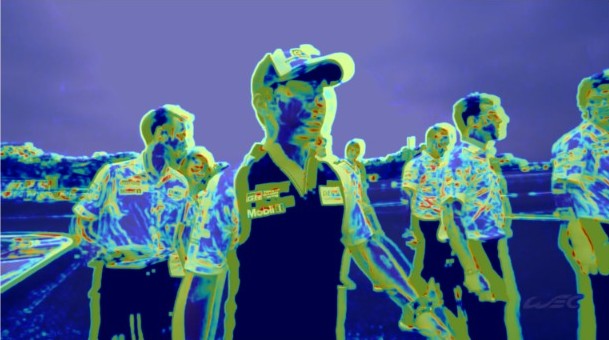} &
\includegraphics[scale=0.13]{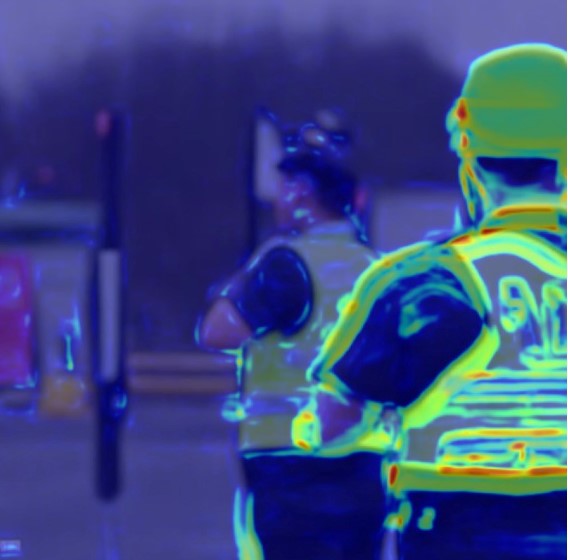} &
\includegraphics[scale=0.13]{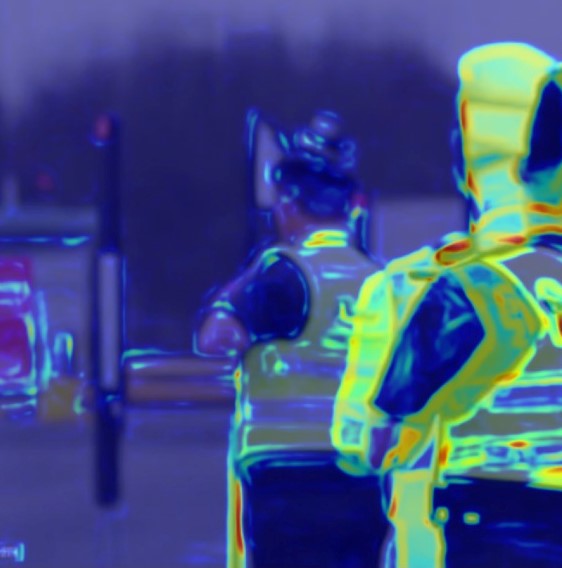} \\
\multicolumn{4}{c}{\scriptsize CLIP appearance attention heatmaps} \\[0.5mm]
\includegraphics[scale=0.13]{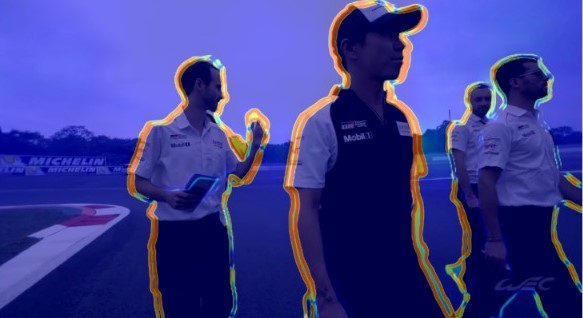} &
\includegraphics[scale=0.13]{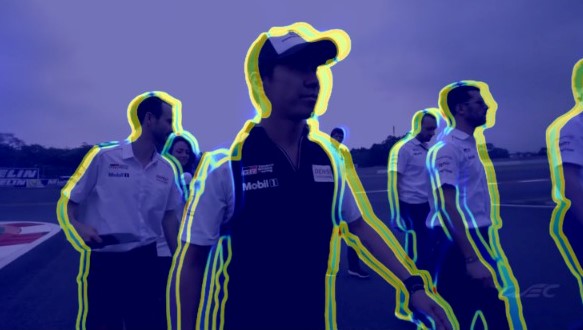} &
\includegraphics[scale=0.13]{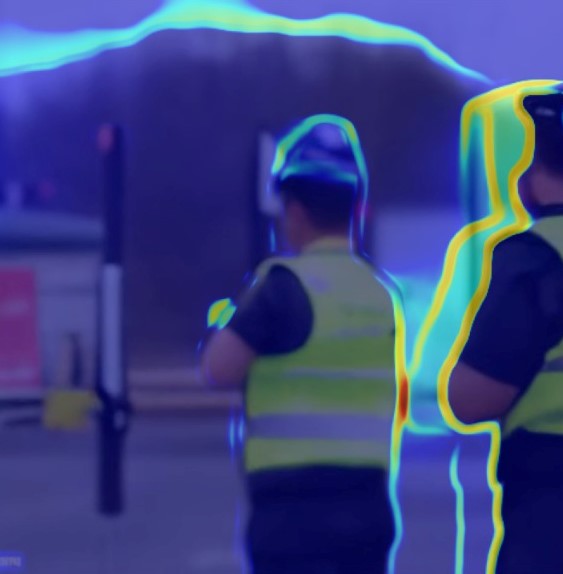} &
\includegraphics[scale=0.13]{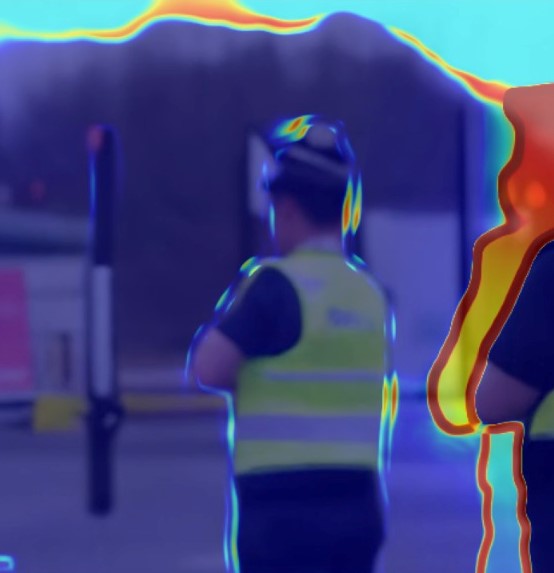} \\
\multicolumn{4}{c}{\scriptsize MiDaS depth attention heatmaps} \\[0.5mm]
\includegraphics[scale=0.13]{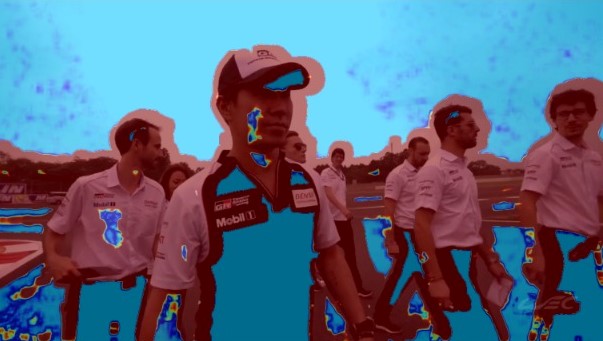} &
\includegraphics[scale=0.13]{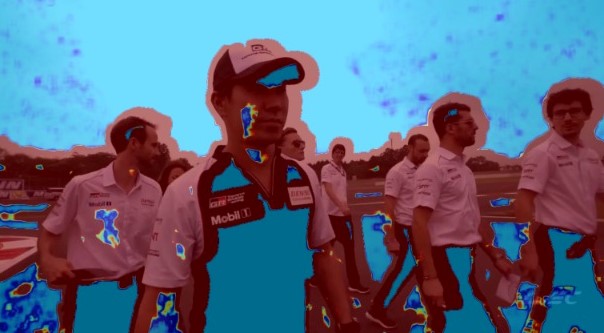} &
\includegraphics[scale=0.13]{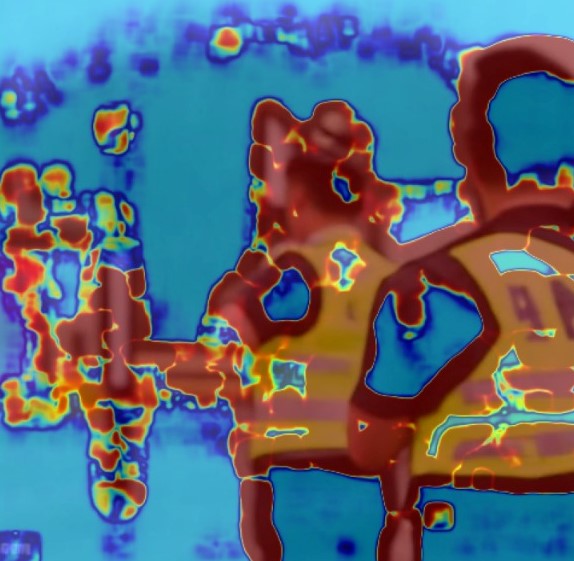} &
\includegraphics[scale=0.13]{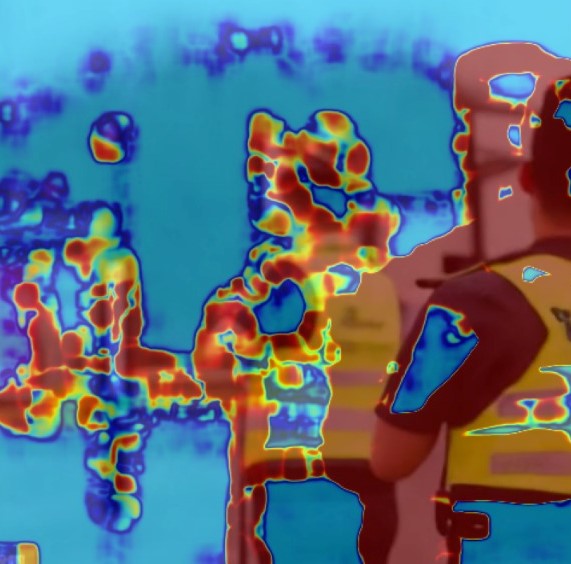} \\
\multicolumn{4}{c}{\scriptsize VideoMAE motion attention heatmaps} \\
\end{tabular}
\vspace{2mm}
\caption{\textbf{Multimodal feature heatmaps extracted from real and
AI-generated video samples by CAM-VFD.} Across all three modalities, real
videos (left) exhibit strong cross-modal consistency, with coherent spatial
structures, stable depth transitions, and physically plausible motion patterns
reflecting underlying scene geometry. In contrast, AI-generated videos (right)
expose distinctive inconsistency signatures: fragmented and spatially unstable
appearance activations, geometrically incoherent depth transitions at subject
boundaries, and abnormal motion energy concentrations at biologically
unrealistic locations.}
\label{fig:heatmap_visual}
\end{figure}

We extract appearance features using the Vision Transformer (ViT-B/32) variant
of the CLIP (Contrastive Language--Image Pre-training) model~\cite{bib17}.
CLIP is a recent vision-language model which captures rich semantic and
structural attributes including facial anatomy, texture realism, and geometric
alignment. As shown in Figure~\ref{fig:heatmap_visual} (row~2), real video
shows coherent and consistent attention patterns across all sampled frames. In
contrast, fake video shows fragmented attention maps, indicating texture and
appearance inconsistencies.

Each input video frame $X_t$ is first resized and normalized using
CLIP-specific parameters ($\mu_{\mathrm{clip}}$, $\sigma_{\mathrm{clip}}$).
Then the appearance feature vector $f_t^{app}$ for each processed frame is
extracted by the CLIP Vision Transformer encoder:
\begin{equation}
f_t^{app} = \mathrm{CLIP\text{-}ViT}\left(\frac{X_t - \mu_{\mathrm{clip}}}{\sigma_{\mathrm{clip}}}\right)
\label{eq:clip_feature}
\end{equation}

To obtain the appearance representation $F_{app}$, we use temporal average
pooling (where $d_{app}$ is the dimensionality of the appearance feature
vector):
\begin{equation}
F_{app} = \frac{1}{T} \sum_{t=1}^{T} f_t^{app} \in \mathbb{R}^{d_{app}}
\label{eq:clip_feature1}
\end{equation}

\subsubsection{Depth Features}

Depth features capture the 3D geometric structure of objects and scenes. Real
videos exhibit consistent depth, whereas AI-generated videos often contain:
\textbf{(i) Geometric Structure Inconsistency:} distorted or unrealistic depth
patterns; \textbf{(ii) Flattened or Collapsed Depth:} inconsistent or
flattened geometric structures, such as unrealistic facial depth; and
\textbf{(iii) Depth Discontinuities at Object Boundaries:} distorted
background due to imperfect 3D rendering. To capture the underlying 3D
geometry and spatial consistency within each frame, we use the MiDaS
(Monocular Depth Estimation in the Wild) DPT-Hybrid model~\cite{bib18} for
computing per-frame depth maps. MiDaS is robust to color variations,
illumination changes, and scene diversity. As shown in
Figure~\ref{fig:heatmap_visual} (row~3), real videos exhibit smooth and
spatially consistent depth distributions across both foreground subjects and
background regions, indicating coherent 3D scene geometry. In contrast, fake
videos expose geometric inconsistencies and unnatural spatial relationships
between the two subjects and the background.

The extraction of depth maps for each frame is performed as follows, where
$D_t$ represents the raw depth map for input frame $X_t$ with a resolution of
$H \times W$ pixels:
\begin{equation}
D_t = \mathrm{MiDaS}(X_t) \in \mathbb{R}^{H \times W}
\label{eq:d_midas}
\end{equation}
Then the depth maps for each input frame are normalized:
\begin{equation}
D_{iz} = \frac{D_t - \mu_D}{\sigma_D + \epsilon}
\label{eq:d_midas1}
\end{equation}
where $D_{iz}$ is the instance-normalized depth map. Subsequently, to
eliminate the effect of global lighting, the normalized depth maps $D_{iz}$
are processed using a lightweight CNN-based encoder (DepthNet: 3 convolutional
layers, max-pool, 128-d output) with batch normalization to obtain depth
embeddings:
\begin{equation}
f_t^{depth} = \mathrm{DepthNet}(\mathrm{BN}(D_{iz}))
\label{eq:d_midas2}
\end{equation}
Here, $BN$ denotes Batch Normalization. Finally, the depth representation is
obtained by averaging across all sampled frames (where $d_{depth}$ is the
dimensionality of the depth feature vector):
\begin{equation}
F_{depth} = \frac{1}{T} \sum_{t=1}^{T} f_t^{depth} \in \mathbb{R}^{d_{depth}}
\label{eq:depth3}
\end{equation}

\subsubsection{Motion Features}

Motion features capture temporal inconsistencies in pixels, objects, or
regions in video frames; they detect:
\textbf{(i) Unsynchronized Motion:} when background and foreground are at
mismatched speeds;
\textbf{(ii) Temporal Flickering:} inconsistent brightness and textures; and
\textbf{(iii) Physics Violations:} unrealistic movement or collision of
objects. We use the VideoMAE (Video Masked Autoencoder) transformer
encoder~\cite{bib20} to capture temporal dynamics glitches. VideoMAE is a
self-supervised pre-trained model designed for video understanding. As shown
in Figure~\ref{fig:heatmap_visual} (row~4), real videos exhibit smoothly and
consistently distributed motion across moving body parts, indicating stable
temporal dynamics. In contrast, fake videos reveal abnormal temporal variation
and motion inconsistency.

Given the adaptively sampled sequence of frames, non-overlapping clips $C_i$
are constructed and fed into the model's dedicated image processor. For each
clip, the motion feature vector $f_i^{motion}$ is extracted from the last
hidden state of the VideoMAE:
\begin{equation}
f_i^{motion} = \mathrm{VideoMAE}(C_i)
\label{eq:motion}
\end{equation}

The global motion embedding $F_{motion}$ is computed as the average across all
clips (where $d_{motion}$ is the dimensionality of the motion feature vector):
\begin{equation}
F_{motion} = \frac{1}{N} \sum_{i=1}^{N} f_i^{motion} \in \mathbb{R}^{d_{motion}}
\label{eq:motion1}
\end{equation}

\subsection{Cross-Modal Fusion}
\label{subsec:fusion}

After extracting the feature representations, it is necessary to combine them
for the \textit{classification} stage. A simple concatenation fusion would
fail because it computes appearance-to-depth alignment and depth-to-appearance
alignment, treating both directions as equally informative, losing the
directional forensic structure. However, symmetric cross-attention improves
upon concatenation but still lacks directional reasoning.

\begin{figure*}[t]
\centering
\includegraphics[scale=0.2]{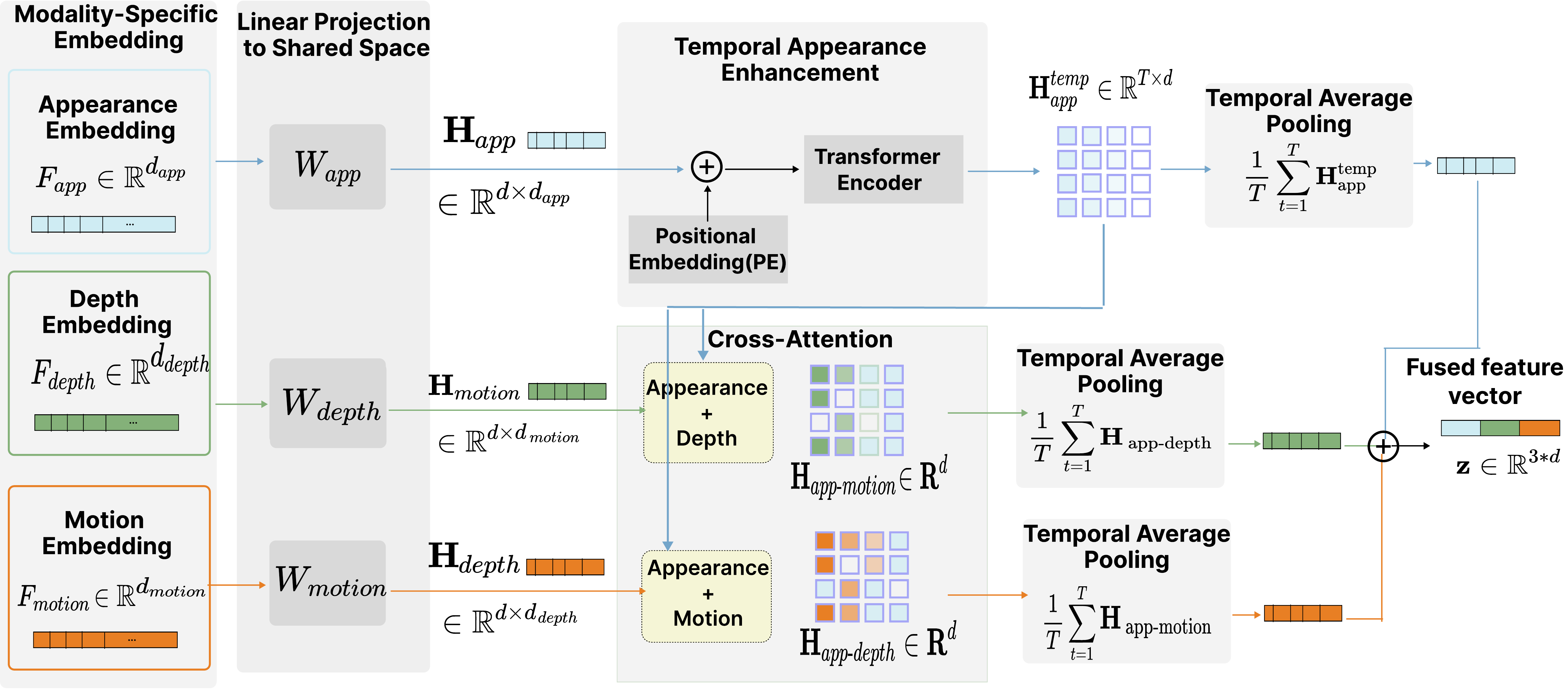}
\caption{Overview of the proposed cross-attention fusion mechanism.}
\label{fig:fusion}
\end{figure*}

Our key insight is that CLIP appearance features encode a rich semantic prior
regarding the expected appearance of a scene. This integration of semantic
content, texture, illumination, and object identity allows appearance to serve
as a natural forensic query channel. By leveraging appearance as queries and
motion/depth as keys and values, the model explicitly poses the question:
\textit{Given the appearance of this scene, do the motion and depth evidence
support or contradict it?} Figure~\ref{fig:fusion} provides a more detailed
illustration of the fusion architecture.

For appearance ($F_{app}$), motion ($F_{motion}$), and depth ($F_{depth}$)
features, we use linear transformations to project them into a shared
dimensional space:
\begin{align}
\mathbf{H}_{app}    &= \mathbf{W}_{app}\mathbf{F}_{app}       \in \mathbb{R}^{d \times d_{app}}    \label{eq:app} \\
\mathbf{H}_{motion} &= \mathbf{W}_{motion}\mathbf{F}_{motion} \in \mathbb{R}^{d \times d_{motion}} \label{eq:mo}  \\
\mathbf{H}_{depth}  &= \mathbf{W}_{depth}\mathbf{F}_{depth}   \in \mathbb{R}^{d \times d_{depth}}  \label{eq:dep}
\end{align}
where $\mathbf{W}_* \in \mathbb{R}^{d \times d_{in}}$ are learnable projection
matrices, $d$ represents the shared latent dimensionality, and $d_{app}$,
$d_{motion}$, and $d_{depth}$ are the input feature dimensionality for the
appearance, motion, and depth modalities, respectively. Projecting all
modalities into a unified dimensional space ensures consistent feature scaling
and allows the cross-attention module to effectively learn inter-modal
dependencies.

To capture inter-frame dependencies before cross-modal querying, the sequence
of appearance embeddings $\mathbf{H}_{app}$ is passed through a Transformer
encoder with learnable positional embedding (PE) to preserve frame order,
yielding temporally-enhanced features $\mathbf{H}_{app}^{temp}$:
\begin{equation}
\mathbf{H}_{app}^{temp} = \mathrm{Transformer}(\mathbf{H}_{app} + \mathbf{PE}) \in \mathbb{R}^{T \times d}
\label{eq:happ}
\end{equation}
where $T$ denotes the number of sampled frames and $d$ represents the shared
latent dimensionality of the temporal transformer encoder. To enable dynamic,
content-aware fusion of multimodal features, we developed two parallel
cross-attention mechanisms. The \textit{first} mechanism computes interactions
between appearance and motion features to identify discrepancies between an
object's visual representation and its movement. The \textit{second} mechanism
evaluates interactions between appearance and depth features to locate areas
where the shape of the object does not correspond with its visual texture. The
computation for the appearance-motion interaction is formalized in
Equation~\ref{eq:appmot}, where the temporally enhanced appearance features
act as the Query ($Q_{app} = H_{app}^{temp}$), while the motion features serve
as both the Key and Value ($K_{motion} = V_{motion} = H_{motion}$). In a
parallel approach, using $H_{depth}$ for the Key and Value calculates the
appearance-depth interaction features $H_{app-depth}$, as shown in
Equation~\ref{eq:appdep}.

\begin{equation}
\mathbf{H}_{app\text{-}motion} = \mathrm{softmax}\!\left(\frac{\mathbf{Q}_{app}\mathbf{K}_{motion}^\top}{\sqrt{d}}\right)\mathbf{V}_{motion}
\label{eq:appmot}
\end{equation}
\begin{equation}
\mathbf{H}_{app\text{-}depth} = \mathrm{softmax}\!\left(\frac{\mathbf{Q}_{app}\mathbf{K}_{depth}^\top}{\sqrt{d}}\right)\mathbf{V}_{depth}
\label{eq:appdep}
\end{equation}

This design means each appearance token explicitly attends to the motion and
depth context: \textit{low} attention weights indicate consistent evidence;
anomalously \textit{high} or structured weights indicate forensic
contradiction. The outputs from each cross-attention block $H_{app-motion}$,
$H_{app-depth}$, and the temporal appearance features $H_{app}^{temp}$ are
pooled by averaging over $T$ (the number of sampled frames per video) and
concatenated to form a unified video representation $z$:
\begin{equation}
z = \left[ \frac{1}{T}\sum_{t=1}^{T} H_{\text{app}}^{\text{temp}};\;\; \frac{1}{T}\sum_{t=1}^{T} H_{\text{app-motion}};\;\; \frac{1}{T}\sum_{t=1}^{T} H_{\text{app-depth}} \right] \in \mathbb{R}^{3d}
\label{eq:mod}
\end{equation}

\subsection{Classification Layer}
\label{subsec:classifier}

The fused representation $z$ is processed through a multi-layer perceptron
(MLP) classifier with two hidden layers; each layer incorporates GELU
activation functions to introduce non-linearity, followed by dropout for
regularization and layer normalization. A final linear layer with a sigmoid
activation function outputs a binary prediction score between 0 and 1. We
classify the video as \emph{Real} if $p \geq 0.5$ and \emph{Fake} otherwise.
The complete prediction pipeline, termed CAM-VFD, is formally defined as:
\begin{equation}
p = \mathrm{CAM\text{-}VFD}(\mathbf{V}) = \mathrm{MLP}(z) \in [0,1]
\label{eq:pred}
\end{equation}

\section{Experiments}
\label{sec:experiments}

This section describes the comprehensive evaluation of CAM-VFD. First, we
introduce the experimental settings. Next, we present the performance of our
model against two benchmark datasets. Then, we conduct ablation studies to
analyze the contribution of different components. Finally, we perform
robustness tests to measure how well our model performs under varying
perturbations.

\subsection{Experimental Settings}
\label{subsec:settings}

\subsubsection{Datasets}

We evaluate CAM-VFD on two complementary benchmarks: GenVidBench~\cite{bib34}
and GenVideo~\cite{bib35}. Together, they provide complementary
generalisation evidence: source-aligned cross-generator robustness vs.\
zero-shot transfer to maximally diverse unseen scenarios.

\textbf{GenVidBench}~\cite{bib34} is a challenging benchmark for AI-generated
video detection, which addresses real-world generalization, semantic diversity,
and cross-generator robustness. It contains 143,400 videos of multiple
resolutions, durations, and generative methods, including 33,931 real videos
and 109,200 synthetic videos generated by eight state-of-the-art
diffusion-based and auto-regressive models using Text-to-Video
(T2V)~\cite{bib12} and Image-to-Video (I2V)~\cite{bib13} frameworks. It is
source-aligned: each synthetic video is generated from the same text prompt or
source image as its paired real video, increasing the difficulty of the
detection task and forcing detectors to rely on subtle forensic artifacts
rather than dataset biases.

\textbf{GenVideo}~\cite{bib35} is a million-scale AI-generated video detection
benchmark comprising over 1,078,838 generated videos and 1,223,511 real
videos. Synthetic videos span 20 distinct state-of-the-art generators and real
videos span 3 diverse sources. The scale and diversity of GenVideo prevent
content-level shortcut exploitation, instead compelling identification of
subtle modality-level authenticity cues.

\subsubsection{Implementation Details}

We fix the temporal dimension to $T=16$ sampled frames employing the adaptive
consecutive sampling strategy, which adjusts the sampling technique according
to the length of the input video, selecting $(6,5,5)$ frames for \textit{long}
videos, two 8-frame segments for \textit{medium-length} videos, and cyclic
repetition for \textit{short} clips. During input preprocessing, video frames
are resized to $256\times256$ and randomly cropped to $224\times224$; the
training set is augmented using \texttt{RandomResizedCrop} (scale
$[0.8,1.0]$), horizontal flipping ($p=0.5$), and \texttt{ColorJitter} with
brightness $0.2$, contrast $0.2$, saturation $0.1$, and hue $0.1$.

For feature extraction, CLIP ViT-B/32 is used for appearance, VideoMAE-Base
for motion, and MiDaS DPT-Hybrid for depth estimation, each producing $512$,
$768$, and $128$ dimensional feature vectors, respectively. All modality
features are projected into a unified $256$-dimensional space. The
cross-attention fusion employs two multi-head attention blocks (8 heads,
$d=256$). The final classifier consists of linear layers of sizes
$768,512,256,1$, using GELU activations, LayerNorm, and dropout rates of
$0.5$ and $0.3$.

All experiments use the Adam optimiser with a learning rate of
$1\times10^{-5}$ and cosine annealing schedule, trained for 50 epochs with
early stopping (patience~$=10$) on a held-out 10\% validation split and batch
size~32. The three pre-trained backbones are frozen during training; only the
projection layers, transformer encoder, cross-attention modules, and MLP
classifier are optimised.

\subsubsection{Evaluation Metrics}

On GenVidBench, we report per-generator accuracy and Top-Mean Accuracy
following the standard evaluation protocol~\cite{bib34}. On GenVideo, we
report Recall, Accuracy, F1-score, and AUROC per generation scenario and
overall average, following the standard evaluation protocol~\cite{bib35}.

\subsection{Results}
\label{subsec:results}

\subsubsection{Comparison on GenVidBench}

Table~\ref{tab:genvidbench} presents a comparative evaluation of our model
against state-of-the-art CNN-based and Transformer-based video classification
models. Training uses videos from Pika, VC2, ModelScope, and Text2Video-Zero
(Group~1) and Vript real videos. Testing uses unseen generators from Group~2
(MuseV, SVD, CogVideo, Mora) and HD-VG-130M. This challenging setup forces
the model to generalize to unseen generation sources. No single-stream
baseline achieves above 83\% mean accuracy, while CAM-VFD exceeds 90\% on all
five subsets, demonstrating strong robustness to unseen generation methods.

CAM-VFD achieves 95.31\% mean Top-1 accuracy, outperforming the best baseline
(MViT V2, 83.27\%) by 12.04\%, while maintaining above 90\% accuracy on every
subset. Accuracy reaches 99.74\% on CogVideo and 99.52\% on HD-VG. Traditional
CNN and Transformer models struggle with generalization as they rely mainly on
appearance cues---for example, MViT V2 achieves 98.29\% on SVD but only
47.50\% on CogVideo. In contrast, CAM-VFD leverages cross-modal contradiction
to capture spatial, temporal, and geometric inconsistencies, supported by a
cross-attention fusion mechanism that enhances robustness across all tested
generators.

\begin{table}[t]
  \centering
  \caption{Comparison with state-of-the-art baselines on GenVidBench.
           Models are trained on Pika, VC2, ModelScope, Text2Video-Zero and
           Vript; tested on unseen sources HD-VG-130M, CogVideo, SVD, MuseV,
           and Mora. Best result in \textbf{bold}.}
  \label{tab:genvidbench}
  \renewcommand{\arraystretch}{1.15}
  \resizebox{\columnwidth}{!}{%
  \begin{tabular}{@{}llcccccr@{}}
    \toprule
    \textbf{Model} & \textbf{Type} & \textbf{HD-VG} & \textbf{CogVideo}
      & \textbf{SVD} & \textbf{MuseV} & \textbf{Mora} & \textbf{Mean} \\
    \midrule
    I3D \cite{bib36}              & CNN   & 93.99 & 60.11 &  8.29 &  8.15 & 59.24 & 45.96 \\
    SlowFast \cite{bib37}         & CNN   & 93.63 & 38.34 & 12.68 & 12.25 & 45.93 & 40.57 \\
    TPN \cite{bib38}              & CNN   & 97.34 & 68.25 &  8.79 & 37.86 & 90.04 & 60.46 \\
    TIN \cite{bib39}              & CNN   & 97.88 & 81.59 & 21.47 & 33.78 & 79.44 & 62.83 \\
    X3D \cite{bib40}              & CNN   & 97.51 & 65.72 & 37.27 & \textbf{92.39} & 49.60 & 68.50 \\
    TRN \cite{bib41}              & CNN   & 93.97 & 91.34 & 26.64 & 38.92 & 93.98 & 68.97 \\
    TSM \cite{bib42}              & CNN   & 96.76 & 78.46 & 54.70 & 70.37 & 70.37 & 74.13 \\
    UniFormerV2 \cite{bib43}      & Xfmr  & 96.89 & 45.21 & 14.81 & 20.05 & \textbf{99.21} & 55.23 \\
    TimeSformer \cite{bib44}      & Xfmr  & 92.32 & 74.80 & 20.17 & 73.14 & 39.40 & 59.97 \\
    VideoSwin \cite{bib45}        & Xfmr  & 98.55 & 88.47 & 68.48 & 49.11 & 81.42 & 77.21 \\
    MViT V2 \cite{bib46}          & Xfmr  & 97.58 & 47.50 & \textbf{98.29} & 76.34 & 96.62 & 83.27 \\
    \midrule
    \textbf{CAM-VFD (ours)} & Multi. & \textbf{99.52} & \textbf{99.74}
      & 92.85 & 92.10 & 92.36 & \textbf{95.31} \\
    \bottomrule
  \end{tabular}}
  \par\vspace{0.5em}
  \footnotesize Xfmr: Transformer. Multi.: Multimodal.
\end{table}

\subsubsection{Comparison on GenVideo}

We evaluate CAM-VFD on the GenVideo benchmark~\cite{bib35} to validate its
generalization beyond the GenVidBench setting. Models are trained on 10,000
real videos from Kinetics-400 and 10,000 generated videos from Pika, then
evaluated on ten generation scenarios entirely unseen during training.
Table~\ref{tab:genvideo} reports Recall, Accuracy, F1-score, and AUROC across
all scenarios and compares CAM-VFD against four competitive baselines.

CAM-VFD achieves 93.43\% average Accuracy (+9.22\% over
DeMamba~\cite{bib35}, +2.19\% over NSG-VD~\cite{bib47}) and 96.56\% AUROC,
maintaining high performance across all ten unseen scenarios. In contrast,
prior methods suffer from severe per-generator degradation; for instance,
STIL~\cite{bib48} collapses to 1.40\% recall on HotShot and 2.00\% on Show1,
while NPR~\cite{bib49} drops to 16.00\% and 33.00\% respectively. CAM-VFD's
worst-case recall drops to 82.30\% on Moon Valley, which contains near-static
scenes with minimal foreground motion that reduces the discriminability of
VideoMAE features. These results confirm that CAM-VFD's cross-modal
contradiction architecture captures fundamental forensic signals, enabling
strong zero-shot generalization.

\begin{table*}[t]
  \centering
  \scriptsize
  \setlength{\tabcolsep}{2pt}
  \renewcommand{\arraystretch}{0.82}
  \caption{Comparison with state-of-the-art baselines on GenVideo.
  Models trained on 10K real (Kinetics-400) and 10K generated (Pika);
  evaluated on unseen generators. Best results in \textbf{bold}.}
  \label{tab:genvideo}
  \resizebox{\textwidth}{!}{%
  \begin{tabular}{@{}llccccccccccc@{}}
  \toprule
    \textbf{Method} & \textbf{Metric}
      & \textbf{Morph} & \textbf{Model} & \textbf{Moon}
      & \textbf{Hot-} & \textbf{Show1} & \textbf{Gen2}
      & \textbf{Craft} & \textbf{LaVie} & \textbf{Sora}
      & \textbf{Wild} & \textbf{Avg.} \\
    & & \textbf{Studio} & \textbf{Scope} & \textbf{Valley}
      & \textbf{Shot} & & & & & & \textbf{Scrape} & \\
    \midrule
    \multirow{4}{*}{DeMamba \cite{bib35}}
      & Recall   & \textbf{87.00} & 93.60 & 98.80 & 40.60 & 48.40 & \textbf{98.00} & 88.40 & 59.00 & 48.21 & 58.20 & 72.02 \\
      & Accuracy & 91.70 & 95.00 & 97.60 & 68.50 & 72.40 & \textbf{97.20} & 92.40 & 77.70 & 72.32 & 77.30 & 84.21 \\
      & F1       & \textbf{91.29} & 94.93 & 97.63 & 56.31 & 63.68 & \textbf{97.22} & 92.08 & 72.57 & 63.53 & 71.94 & 80.12 \\
      & AUROC    & 98.04 & \textbf{98.82} & 99.68 & 87.84 & 90.12 & 99.46 & 97.81 & 91.32 & 88.36 & 87.38 & 93.88 \\
    \midrule
    \multirow{4}{*}{NPR \cite{bib49}}
      & Recall   & 61.20 & 80.00 & 98.00 & 16.00 & 33.00 & 91.20 & 80.60 & 34.60 & 35.71 & 43.20 & 57.35 \\
      & Accuracy & 79.80 & 89.20 & \textbf{98.20} & 57.20 & 65.70 & 94.80 & 89.50 & 66.50 & 67.86 & 70.80 & 77.96 \\
      & F1       & 75.18 & 88.11 & \textbf{98.20} & 27.21 & 49.03 & 94.61 & 88.47 & 50.81 & 52.63 & 59.67 & 68.39 \\
      & AUROC    & 93.05 & 97.18 & 99.66 & 82.97 & 90.50 & 99.13 & 97.87 & 87.54 & 90.47 & 91.84 & 93.02 \\
    \midrule
    \multirow{4}{*}{TALL \cite{bib50}}
      & Recall   & 51.20 & 65.20 & 93.40 & 32.00 & 61.60 & 94.80 & 81.80 & 49.20 & 25.00 & 53.60 & 60.78 \\
      & Accuracy & 75.10 & 82.10 & 96.20 & 65.50 & 80.30 & 96.90 & 90.40 & 74.10 & 61.61 & 76.30 & 79.85 \\
      & F1       & 67.28 & 78.46 & 96.09 & 48.12 & 75.77 & 96.83 & 89.50 & 65.51 & 39.44 & 69.34 & 72.63 \\
      & AUROC    & 95.82 & 97.14 & \textbf{99.73} & 92.55 & \textbf{97.36} & \textbf{99.79} & \textbf{99.09} & \textbf{94.84} & 86.67 & \textbf{93.75} & \textbf{95.67} \\
    \midrule
    \multirow{4}{*}{STIL \cite{bib48}}
      & Recall   & 73.80 & 70.80 & 43.40 & 1.40 & 2.00 & 45.00 & 13.20 & 7.20 & 1.79 & 11.60 & 27.02 \\
      & Accuracy & 86.90 & 85.40 & 71.70 & 50.70 & 51.00 & 72.50 & 56.60 & 53.60 & 50.89 & 55.80 & 63.51 \\
      & F1       & 84.93 & 82.90 & 60.53 & 2.76 & 3.92 & 62.07 & 23.32 & 13.43 & 3.51 & 20.79 & 35.82 \\
      & AUROC    & 96.43 & 97.77 & 99.34 & 86.66 & 90.56 & 98.88 & 97.04 & 88.16 & 92.57 & 87.52 & 93.49 \\
    \midrule
    \multirow{4}{*}{NSG-VD \cite{bib47}}
      & Recall   & 68.33 & \textbf{98.33} & \textbf{100.00} & \textbf{92.50} & 87.50 & 80.00 & \textbf{98.33} & \textbf{94.17} & 78.57 & 82.50 & \textbf{88.02} \\
      & Accuracy & 81.67 & \textbf{98.33} & 96.67 & 91.67 & 90.83 & 88.33 & 95.83 & \textbf{94.17} & 88.39 & 88.75 & 91.46 \\
      & F1       & 78.85 & \textbf{98.33} & 96.77 & 91.74 & 90.52 & 87.27 & 95.93 & \textbf{94.17} & \textbf{87.13} & 88.00 & 90.87 \\
      & AUROC    & 92.26 & 98.66 & 98.15 & \textbf{94.45} & 96.38 & 94.83 & 98.16 & 97.41 & \textbf{96.40} & 94.73 & \textbf{96.14} \\
    \midrule
    \multirow{4}{*}{\textbf{CAM-VFD (ours)}}
      & Recall   & 85.20 & 89.50 & 82.30 & 91.80 & \textbf{90.50} & 86.70 & 93.20 & 88.40 & \textbf{83.60} & \textbf{87.90} & 87.91 \\
      & Accuracy & \textbf{92.50} & 94.80 & 90.60 & \textbf{95.90} & \textbf{95.10} & 93.20 & \textbf{96.80} & 93.60 & \textbf{89.50} & \textbf{92.30} & \textbf{93.43} \\
      & F1       & 88.70 & 92.10 & 86.40 & \textbf{93.80} & \textbf{92.80} & 89.90 & \textbf{95.00} & 91.00 & 86.50 & \textbf{90.10} & \textbf{90.63} \\
      & AUROC    & 95.80 & 97.20 & 94.90 & \textbf{98.10} & \textbf{97.60} & 96.40 & 98.70 & 96.90 & 93.80 & \textbf{96.20} & \textbf{96.56} \\
    \bottomrule
  \end{tabular}%
  }
\end{table*}

\subsection{Generalization Evaluation}

To examine our method's generalizability, Table~\ref{tab:generalization}
reports performance under In-Distribution (ID) and Out-of-Distribution (OOD)
evaluation on GenVidBench. ID means testing on the same generator seen during
training; OOD means testing on a different unseen generator. In Group~1 (Pika,
VideoCrafter2, ModelScope, and Text2Video-Zero), all videos share identical
text prompts. In Group~2, videos are generated from either the same image
(MuseV, SVD) or the same text prompt (CogVideo, Mora).

The ID-to-OOD accuracy drop for CAM-VFD is minimal: from 99.20\% to 97.52\%
for Group~1 (mean drop of 1.68\%) and from 96.61\% to 93.98\% for Group~2
(mean drop of 2.63\%). By comparison, VideoSwin-Tiny achieves only 56.71\%
OOD on Group~2 under equivalent conditions~\cite{bib34}. These results
highlight that cross-modal contradiction is a generator-agnostic forensic
signal that does not overfit to the artifact characteristics of any specific
generation method.

\begin{table}[t]
  \centering
  \caption{Cross-generator generalization performance across in-distribution
           (ID) and out-of-distribution (OOD) generators.}
  \label{tab:generalization}
  \renewcommand{\arraystretch}{1.15}
  \begin{tabular}{@{}lcc@{}}
    \toprule
    \textbf{Video Source} & \textbf{ID Acc.\ (\%)} & \textbf{OOD Acc.\ (\%)} \\
    \midrule
    \multicolumn{3}{@{}l}{\textit{Group 1: Text-to-Video Generators}} \\
    \quad Pika             & 98.02 & 97.37 \\
    \quad VideoCrafter2    & 99.31 & 98.05 \\
    \quad Text2Video-Zero  & 99.55 & 95.66 \\
    \quad ModelScope       & 99.91 & 99.00 \\
    \quad\textbf{Mean}     & \textbf{99.20} & \textbf{97.52} \\
    \midrule
    \multicolumn{3}{@{}l}{\textit{Group 2: Image/Text-to-Video \& High-Quality Fakes}} \\
    \quad SVD              & 92.97 & 89.71 \\
    \quad MuseV            & 95.94 & 92.15 \\
    \quad Mora             & 97.63 & 94.27 \\
    \quad CogVideo         & 99.89 & 99.80 \\
    \quad\textbf{Mean}     & \textbf{96.61} & \textbf{93.98} \\
    \bottomrule
  \end{tabular}
\end{table}

\subsection{Ablation Study}
\label{sec5}

All ablation studies isolate one component while holding the others fixed,
compared to the full model (three modalities, 16 frames, cross-attention
fusion) trained and tested on ModelScope and Vript, achieving 99.91\%
accuracy.

\subsubsection{Modality Ablation}

Table~\ref{tab:ablation_modality} shows an accuracy comparison between the
full model and multiple modality configurations, proving the impact of
different input modalities. \textit{Appearance-only} achieves 86.0\%,
confirming that visual semantics provides a useful baseline. However,
\textit{motion} features achieve 93.0\%, highlighting that temporal dynamics
are more discriminative than static appearance. \textit{Depth}-only achieves
82.0\%, showing that geometric cues are most valuable in combination. The
combination of modalities improves accuracy substantially: appearance and
motion reaches 98.0\%, appearance and depth reaches 97.0\%, and motion and
depth achieves 95.0\%. The full three-modality model with cross-attention
fusion achieves 99.91\%, confirming that each modality provides complementary
forensic evidence.

\begin{table}[t]
  \centering
  \caption{Modality contribution ablation.}
  \label{tab:ablation_modality}
  \renewcommand{\arraystretch}{1.15}
  \begin{tabular}{@{}lc@{}}
    \toprule
    \textbf{Configuration} & \textbf{Acc.\ (\%)} \\
    \midrule
    Appearance only          & 86.00 \\
    Motion only              & 93.00 \\
    Depth only               & 82.00 \\
    Appearance + Motion      & 98.00 \\
    Appearance + Depth       & 97.00 \\
    Motion + Depth           & 95.00 \\
    \midrule
    \textbf{All modalities (proposed)} & \textbf{99.91} \\
    \bottomrule
  \end{tabular}
\end{table}

\subsubsection{Temporal Ablation}

Table~\ref{tab:ablation_temporal} explores the effects of varying the number
of input frames, fixing the model to use all modalities and attention fusion.
Using only 8 frames achieves an accuracy of 90\%, indicating insufficient
temporal coverage. Increasing to 16 frames improves accuracy to 99.91\%. A
32-frame configuration reaches 99.95\%, confirming that a longer temporal
window allows the model to learn more discriminative features. However, the
marginal gain from 16 to 32 frames is only $+0.04\%$, revealing that doubling
the number of frames yields only a minimal gain in accuracy at double the
computational cost. Therefore, 16 frames are selected as the optimal temporal
balance.

\begin{table}[t]
  \centering
  \caption{Temporal window ablation.}
  \label{tab:ablation_temporal}
  \renewcommand{\arraystretch}{1.15}
  \begin{tabular}{@{}lc@{}}
    \toprule
    \textbf{Frame Count} & \textbf{Acc.\ (\%)} \\
    \midrule
    8  frames                         & 90.00 \\
    \midrule
    \textbf{16 frames (selected)}     & \textbf{99.91} \\
    \midrule
    32 frames                         & 99.95 \\
    \bottomrule
  \end{tabular}
\end{table}

\subsubsection{Fusion Strategy and Query Direction Ablation}

Table~\ref{tab:ablation_fusion} evaluates various architectural strategies for
cross-modal integration and validates the asymmetric design direction. Simple
concatenation achieves 90\%, confirming that naive feature stacking fails to
capture inter-modal dependencies. Late fusion, which aggregates predictions
after each stream, improves results to 97.5\%. Symmetric cross-attention
further improves accuracy to 98.40\%, demonstrating the importance of feature
interaction but highlighting the limitation of treating modalities equivalently
without directional structure. To prove that modeling directional relationships
is critical, we evaluate motion-as-query and depth-as-query, achieving 97.23\%
and 96.88\% respectively. The proposed cross-attention fusion (appearance as
query, motion and depth as keys/values) achieves 99.91\%. These results
demonstrate that the performance gains of CAM-VFD arise from structured
cross-modal reasoning rather than simple multi-stream aggregation.

\begin{table}[t]
  \centering
  \caption{Fusion strategy and query direction ablation.}
  \label{tab:ablation_fusion}
  \renewcommand{\arraystretch}{1.15}
  \begin{tabular}{@{}lc@{}}
    \toprule
    \textbf{Fusion Method} & \textbf{Acc.\ (\%)} \\
    \midrule
    Concatenation                 & 90.00 \\
    Late fusion                   & 97.50 \\
    Symmetric cross-attention     & 98.40 \\
    Motion queries, App.\ K/V    & 97.23 \\
    Depth queries, App.\ K/V     & 96.88 \\
    \midrule
    \textbf{App.\ queries (proposed)} & \textbf{99.91} \\
    \bottomrule
  \end{tabular}
\end{table}

\section{Cross-Modal Attention Discrepancy Analysis}
\label{sec:discrepancy}

CAM-VFD is grounded in a core forensic insight: real videos are physically
consistent across appearance, geometry, and motion, as they arise from the
same real-world scene. In contrast, AI-generated videos are produced through
pipelines whose components lack this physical consistency constraint, producing
cross-modal contradictions that are detectable even when each modality appears
individually plausible.

Given a video $V = \{x_1, \dots, x_T\}$, we define the Cross-Modal Attention
Discrepancy (CMAD) as:
\begin{equation}
\begin{aligned}
\text{CMAD}(V) = \frac{1}{2T} \sum_{t=1}^{T} \Big(
&\left\| \mathbf{H}_{\text{app-motion}}^{(t)} - \mathbf{H}_{\text{app}}^{temp,(t)} \right\|_2^2 \\
&+ \left\| \mathbf{H}_{\text{app-depth}}^{(t)} - \mathbf{H}_{\text{app}}^{temp,(t)} \right\|_2^2
\Big)
\end{aligned}
\end{equation}
where $\mathbf{H}_{app}^{temp}$ is the temporal appearance features and
$\mathbf{H}_{\text{app-motion}}$, $\mathbf{H}_{\text{app-depth}}$ are the
cross-attended representations after querying motion and depth respectively.
CMAD measures the degree of cross-modal inconsistency by quantifying how much
the motion and depth evidence contradict the appearance evidence. Under this
formulation, real videos are expected to produce low CMAD values, whereas fake
videos produce higher CMAD values.

\begin{figure}[t]
\centering
\includegraphics[scale=0.8]{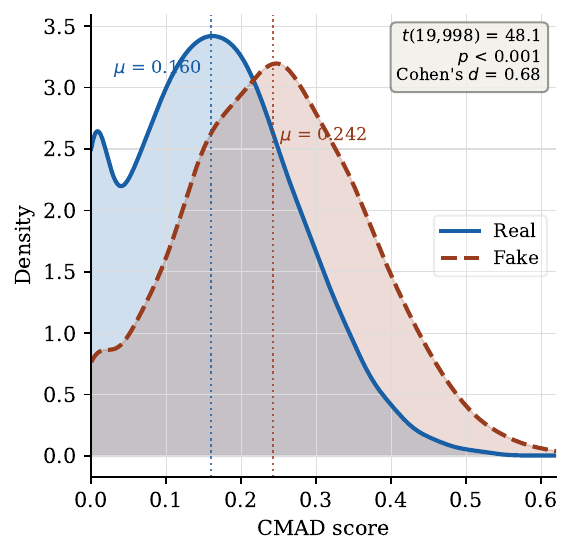}
\caption{Distribution of Cross-Modal Attention Discrepancy (CMAD) scores for
real and fake videos on the GenVideo benchmark.}
\label{fig:cmad_analysis}
\end{figure}

To validate that CMAD separates real from fake videos, we compute CMAD across
the GenVideo benchmark. As shown in Figure~\ref{fig:cmad_analysis}, real
videos exhibit consistently low discrepancy values, reflecting coherent
cross-modal alignment consistent with physical reality. In contrast, fake
videos produce significantly higher and more localized discrepancy patterns
corresponding to cross-modal contradictions.

To assess statistical significance, we perform a two-sample $t$-test on the
CMAD distributions of real and fake videos~\cite{bib51} as shown in
Table~\ref{tab:cmad_analysis}. The test yields a highly significant separation
$p < 0.001$, confirming that the observed difference is not due to random
variation. We further report Cohen's $d$ as an effect size
measure~\cite{bib52}, indicating a large effect size ($d = 0.68$) between the
two distributions. These results confirm that the cross-modal contradiction
signal is strongly discriminative and is consistent with the qualitative
feature heatmap visualisations in Figure~\ref{fig:heatmap_visual}, which
visibly demonstrate cross-modal incoherence in fake videos across all three
modality branches.

\begin{table}[t]
\centering
\caption{Cross-Modal Attention Discrepancy (CMAD) statistical analysis on
GenVideo.}
\label{tab:cmad_analysis}
\setlength{\tabcolsep}{7pt}
\renewcommand{\arraystretch}{1.15}
\begin{tabular}{lcc}
\toprule
\textbf{Statistic} & \textbf{Real Videos} & \textbf{Fake Videos} \\
\midrule
Sample size ($n$)           & 10,000 & 10,000 \\
Mean CMAD ($\mu$)           & 0.160  & 0.242  \\
Std.\ deviation ($\sigma$)  & 0.115  & 0.125  \\
\midrule
\multicolumn{3}{l}{\textit{Welch two-sample $t$-test (real vs.\ fake)}} \\
\midrule
$t$-statistic ($df = 19{,}998$) & \multicolumn{2}{c}{48.1}      \\
$p$-value                        & \multicolumn{2}{c}{$<0.001$} \\
Cohen's $d$                      & \multicolumn{2}{c}{0.68}      \\
Pooled SD                        & \multicolumn{2}{c}{0.120}     \\
\bottomrule
\end{tabular}
\end{table}

\section{Robustness Evaluation}
\label{sec:robustness}

We evaluated the model's robustness under real-world conditions across two
complementary threat categories: passive signal degradation
(Section~\ref{ssec:rob_signal}) and active adversarial attack
(Section~\ref{ssec:rob_adversarial}). All experiments are conducted on two
representative GenVidBench subsets: CogVideo (highest accuracy, 99.68\%) and
SVD (lowest accuracy, 90.22\%). DeMamba~\cite{bib35} is evaluated under
identical corruption conditions throughout Table~\ref{tab:rob_signal}.

\subsection{Signal Degradation Robustness}
\label{ssec:rob_signal}

Table~\ref{tab:rob_signal} presents accuracy under three categories of passive
signal degradation: video compression (Constant Rate Factor CRF $\in
\{18,28,32\}$, where higher CRF indicates heavier compression), additive noise
(Gaussian $\sigma \in \{0.03, 0.1\}$ and salt-and-pepper $p \in \{0.01,
0.05\}$), and spatial blur (Gaussian $\sigma \in \{1,2\}$, defocus radius
$r \in \{3,7\}$, and motion kernel $L \in \{7,21\}$). Each condition is
applied independently; all other processing is identical to the standard
inference pipeline.

Across all conditions and both subsets, CAM-VFD outperforms DeMamba by a
consistent margin. Under heavier corruption at CRF$=32$, CAM-VFD leads
DeMamba by 14.9\% on SVD and 11.4\% on CogVideo, achieving 83.9\% and 91.6\%
accuracy while DeMamba achieves 69.0\% and 80.2\% respectively. CAM-VFD drops
most under strong motion blur ($L{=}21$), where temporal cues are most
directly damaged: SVD falls to 76.7\% and CogVideo to 86.0\%. This is the
expected worst case because VideoMAE motion features are most sensitive to
motion blur distortion, while CLIP and MiDaS features, which operate on
individual frames, provide partial but incomplete compensation. Noise and
lighting distortions produce the smallest degradation, reflecting the
pre-trained backbones' focus on abstract semantic representations that are
inherently insensitive to high-frequency or photometric perturbations.

CAM-VFD similarly exhibits negligible accuracy change under photometric
perturbations (Table~\ref{tab:photometric}): brightness offsets
($\Delta \leq 0.1$), contrast scaling ($0.7 \leq \alpha \leq 1.3$),
saturation shifts ($0.7 \leq s \leq 1.3$), and hue rotation
($\Delta h \leq 12^\circ$), with all photometric conditions maintaining
accuracy above 86\% on both subsets.

\begin{table*}[t]
  \centering
  \caption{Signal degradation robustness: accuracy (\%) of CAM-VFD and
           DeMamba~\cite{bib35} under identical compression, noise, and blur
           conditions on two GenVidBench subsets (CogVideo and SVD).}
  \label{tab:rob_signal}
  \renewcommand{\arraystretch}{1.2}
  \resizebox{\textwidth}{!}{%
  \begin{tabular}{@{}ll
      c
      ccc
      cccc
      cccccc
    @{}}
    \toprule
    & &
    &
    \multicolumn{3}{c}{\textbf{Compression (H.264 CRF)}} &
    \multicolumn{4}{c}{\textbf{Noise}} &
    \multicolumn{6}{c}{\textbf{Blur}} \\
    \cmidrule(lr){4-6}\cmidrule(lr){7-10}\cmidrule(lr){11-16}
    & & &
    & & &
    \multicolumn{2}{c}{\textit{Gaussian}} &
    \multicolumn{2}{c}{\textit{Salt-Pepper}} &
    \multicolumn{2}{c}{\textit{Gaussian}} &
    \multicolumn{2}{c}{\textit{Defocus}} &
    \multicolumn{2}{c}{\textit{Motion}} \\
    \cmidrule(lr){7-8}\cmidrule(lr){9-10}
    \cmidrule(lr){11-12}\cmidrule(lr){13-14}\cmidrule(lr){15-16}
    \textbf{Subset} & \textbf{Method} &
    \textbf{Base} &
    \textbf{18} & \textbf{28} & \textbf{32} &
    $\sigma{=}0.03$ & $\sigma{=}0.1$ &
    $p{=}0.01$ & $p{=}0.05$ &
    $\sigma{=}1$ & $\sigma{=}2$ &
    $r{=}3$ & $r{=}7$ &
    $L{=}7$ & $L{=}21$ \\
    \midrule
    \multirow{2}{*}{CogVideo}
      & DeMamba~\cite{bib35}
        & 97.6 & 96.1 & 89.7 & 80.2
        & 93.8  & 82.4  & 91.0  & 84.6
        & 88.3  & 79.5  & 84.1  & 80.9  & 83.5  & 74.8  \\
      & \textbf{CAM-VFD}
        & \textbf{99.7} & \textbf{99.2} & \textbf{96.8} & \textbf{91.6}
        & \textbf{98.0} & \textbf{90.8} & \textbf{95.1} & \textbf{91.4}
        & \textbf{95.9} & \textbf{98.5} & \textbf{93.2} & \textbf{90.7}
        & \textbf{91.8} & \textbf{86.0} \\
    \midrule
    \multirow{2}{*}{SVD}
      & DeMamba~\cite{bib35}
        & 84.3 & 82.1 & 77.8 & 69.0
        & 78.2  & 73.5  & 76.9  & 70.4
        & 74.1  & 68.3  & 72.6  & 65.8  & 70.2  & 61.4  \\
      & \textbf{CAM-VFD}
        & \textbf{90.2} & \textbf{89.8} & \textbf{88.4} & \textbf{83.9}
        & \textbf{89.3} & \textbf{85.7} & \textbf{88.4} & \textbf{81.2}
        & \textbf{87.5} & \textbf{81.2} & \textbf{85.7} & \textbf{78.5}
        & \textbf{83.0} & \textbf{76.7} \\
    \bottomrule
  \end{tabular}}
\end{table*}

\begin{table*}[t]
\centering
\small
\caption{Robustness of CAM-VFD under photometric perturbations, including
lighting (brightness, contrast) and color distortions (saturation, hue
shift).}
\label{tab:photometric}
\renewcommand{\arraystretch}{1.1}
\begin{tabular*}{\textwidth}{@{\extracolsep{\fill}}lccccccccc}
\toprule
& \multicolumn{4}{c}{\textbf{Lighting}}
& \multicolumn{4}{c}{\textbf{Color Distortion}} \\
\cmidrule(lr){2-5} \cmidrule(lr){6-9}
\textbf{Subset}
& \textbf{Base}
& $\Delta\!=\!0.05$
& $\Delta\!=\!0.1$
& $\alpha\!=\!0.7$
& $\alpha\!=\!1.3$
& $s\!=\!0.7$
& $s\!=\!1.3$
& $\Delta h\!=\!5^\circ$
& $\Delta h\!=\!12^\circ$ \\
\midrule
CogVideo & 99.7 & 99.7 & 98.3 & 96.0 & 98.4 & 97.1 & 98.5 & 97.5 & 93.2 \\
SVD      & 90.2 & 90.2 & 89.8 & 87.5 & 86.6 & 89.0 & 88.4 & 88.4 & 86.6 \\
\bottomrule
\end{tabular*}
\end{table*}

\subsection{Adversarial Robustness}
\label{ssec:rob_adversarial}

We evaluate the robustness of CAM-VFD under standard adversarial threat models
using FGSM and PGD-20 attacks~\cite{bib53}. We apply FGSM (single-step) and
PGD-20 (iterative) adversarial attacks with $\epsilon \in \{2/255, 4/255,
8/255\}$ defining the maximum perturbation magnitude applied to each pixel. As
reported in Table~\ref{tab:adversarial}, CAM-VFD exhibits relatively
consistent robustness across increasing perturbation strengths, demonstrating
strong robustness across both single-stream (appearance-only) and full-model
attack settings. Under appearance attacks, performance decreases gradually from
$99.74\%$ in the clean setting to 84.17\% under FGSM and 79.43\% under the
strongest PGD-20 attack. This is a direct consequence of the multimodal
architecture: perturbing the appearance stream does not affect VideoMAE or
MiDaS. Under the full-model attack, where perturbations are applied across all
modalities, performance degradation remains moderate (82.31\% and 77.68\%).
These results confirm the robustness of the proposed cross-attention fusion
mechanism to both feature-level and joint multimodal attacks.

\begin{table}[t]
  \centering
  \caption{Adversarial robustness under FGSM and PGD-20 attacks on the
           CogVideo subset of GenVidBench (clean accuracy: 99.74\%).}
  \label{tab:adversarial}
  \renewcommand{\arraystretch}{1.15}
  \begin{tabular}{@{}lccc@{}}
    \toprule
    \textbf{Attack} & $\boldsymbol{\epsilon}$
      & \textbf{App.-only (\%)}
      & \textbf{Full-model (\%)} \\
    \midrule
    FGSM   & $2/255$ & 96.44 & 95.22 \\
    FGSM   & $4/255$ & 91.32 & 88.94 \\
    FGSM   & $8/255$ & 84.17 & 82.31 \\
    \midrule
    PGD-20 & $4/255$ & 88.66 & 85.42 \\
    PGD-20 & $8/255$ & 79.43 & 77.68 \\
    \bottomrule
  \end{tabular}
\end{table}

\section{Conclusion}
\label{sec:conclusion}

Going beyond traditional facial manipulations to fabricate entire video scenes
using AI-powered generation tools represents a direct and escalating threat to
digital security: to the integrity of forensic evidence, to the authenticity
of information, and to the trustworthiness of digital identity. Addressing
this threat demands detection systems that are robust and generalizable. In
this paper, we introduced \textit{CAM-VFD}, a Cross-Attention Multimodal
Video Forgery Detection framework for detecting fully AI-generated videos. The
approach is based on the principle that cross-modal contradiction---systematic
misalignment between appearance, geometric depth, and motion dynamics---is a
more stable and generalisable forgery signal than any within-modality
artifact. This design improves generalization to unseen generative models and
enhances robustness under real-world distortions, including compression,
noise, blur, and adversarial perturbations. Furthermore, our proposed
Cross-Modal Attention Discrepancy (CMAD) analysis provides interpretability by
quantifying the cross-modal inconsistency between real and fake videos.

Despite these advantages, CAM-VFD still exhibits limitations in scenarios
where motion cues are weak or temporally ambiguous, such as near-static videos
or extremely low-motion scenes, which reduce the discriminability of VideoMAE
features. Moreover, the rapid evolution of synthetic video generators may pose
future constraints. The proposed framework establishes multimodal
cross-attention fusion as a dependable forensic strategy for AI-generated video
detection, and provides a strong foundation for future work in which we aim to
further explore adaptive modality and frame sampling to improve performance and
robustness. Finally, integrating explainable AI (XAI) techniques could provide
interpretable forensic evidence for model decisions, increasing transparency
and trust in forensic verification systems.

\bibliographystyle{IEEEtran}
\bibliography{references}

\end{document}